# Exploiting the Rule Structure for Decision Making within the Independent Choice Logic

David Poole*
Department of Computer Science
University of British Columbia
Vancouver, B.C., Canada V6T 1Z4
poole@cs.ubc.ca
http://www.cs.ubc.ca/spider/poole

## Abstract

This paper introduces the independent choice logic, and in particular the "single agent with nature" instance of the independent choice logic, namely $ICL_{DT}$. This is a logical framework for decision making uncertainty that extends both logic programming and stochastic models such as influence diagrams. This paper shows how the representation of a decision problem within the independent choice logic can be exploited to cut down the combinatorics of dynamic programming. One of the main problems with influence diagram evaluation techniques is the need to optimise a decision for all values of the 'parents' of a decision variable. In this paper we show how the rule based nature of the $ICL_{DT}$ can be exploited so that we only make distinctions in the values of the information available for a decision that will make a difference to utility.

## 1 Introduction

Most current approaches to solving decision problems under uncertainty involve a case analysis on all available information (for example on all current and past observations and past actions in influence diagrams [Howard and Matheson, 1981; Cooper, 1988; Shachter and Peot, 1992; Qi and Poole, 1995], or on the current belief state in partially observable Markov decision problems (POMDPs) [Monahan, 1982; Cassandra et al., 1994]).

In this paper, we consider how a logic-based representation of decision problems that treats causal rules as logic programs can be exploited to reduce the case analysis for dynamic programming. This representation that allows one to express logical rules and choices made by various agents, is capable of representing general decision problems (that extends both influence diagrams and (finite stage) POMDPs).

The logic is the independent choice logic (ICL) that allows for a space of independent choices and a logic program that gives the consequences of these choices. The choices can be made by nature (which has probabilities over the choices) or by purposive agents (who are trying to maximise their utility). The ICL extends the author's probabilistic Horn abduction [Poole, 1993b] to include negation as failure and multiple agents. In this paper we only consider the decision theoretic (single agent under uncertainty) case. For the no-agent case (with probabilities over choices), the rules induce an independence equivalent to that of Bayesian networks. The rules also allow the representation of a form of 'propositional independence' where one variable may only be dependent on another for some values of a third variable. It is this last property that we exploit in this paper.

The main point of this paper is to show how the rule-structure can be exploited to gain efficiency. The rules provide a modular specification of utility, and a modular specification of what will be observed when a decision is made (this is similar to using decision trees to specify the probability and utility tables [Smith et al., 1993; Boutilier et al., 1995]). Instead of optimizing a decision for each of its information states, we 'mesh' the decision trees for the 'observables' (the information available when the decision is made) and the decision trees for the utilities, and only make the distinctions in the observables that matter (would lead to different utilities). We show by example that this can cut down on the number of optimizations that we need to do. The meshing becomes complicated when interleaved with dominance testing — we want to prune dominated decisions as soon as possible, so we don't make distinctions that are only important for decisions than can be shown to be non-optimal.

This paper could have been described in terms of decision trees (as does [Boutilier et al., 1995] using a similar idea for fully observable MDPs, see Section 6). This was not done for a number of reasons. The ICL forms a simple logical framework that includes influence diagrams (the rules can encode all of the dependencies of an influence diagram

---

— we don't need the influence diagram *and* the rules), as well as being interesting in its own right as a mix of logic and decision/game theory [Poole, 1995b]. The meshing is also easily described in this framework in terms of 'explanations'. The ICL also naturally has a way to include logical variables, and thus we allow for parametrizable influence diagrams (see [Poole, 1993b] for a description of the purely probabilistic case).

## 2 The Independent Choice Logic

The Independent Choice Logic specifies a way to build possible worlds. Possible worlds are built from choosing propositions from independent alternatives, and then extending these 'total choices' with a logic program. This section defines the single agent case ICL$_{DT}$.

There are two languages we will use: $\mathcal{L}_F$ which, for this paper, is the language of acyclic logic programs [Apt and Bezem, 1991], and the language $\mathcal{L}_Q$ of queries which we take to be arbitrary propositional formulae (the atoms corresponding to ground atomic formulae of the language $\mathcal{L}_F$). We write $f \hspace{1pt}\vert\!\sim q$ where $f \in \mathcal{L}_F$ and $q \in \mathcal{L}_Q$ if $q$ is true in the unique stable model of $f$ or, equivalently, if $q$ follows from Clark's completion of $q$ (the uniqueness of the stable model and the equivalence for acyclic programs are proved in [Apt and Bezem, 1991]). See [Poole, 1995a] for a detailed analysis of negation as failure in this framework, and for an abductive characterisation of the logic.

**Definition 2.1** A **choice space** is a set of sets of ground atomic formulae, such that if $C_1$, and $C_2$ are in the choice space, and $C_1 \neq C_2$ then $C_1 \cap C_2 = \{\}$. An element of a choice space is called a **choice alternative** (or sometimes just an alternative). An element of a choice alternative is called an **atomic choice**. An atomic choice can appear in at most one alternative.[1]

**Definition 2.2** An ICL$_{DT}$ theory is a tuple $\langle \mathcal{C}_0, \mathcal{C}_1, \mathcal{O}, \pi, P_0, \mathcal{F} \rangle$ where

$\mathcal{C}_0$ called **nature's choice space**, is the choice space of alternatives controlled by nature.

$\mathcal{C}_1$ called the **agent's choice space**, is the choice space of alternatives controlled by our agent (what the agent decides to do). No atomic choice can be both in an element of $\mathcal{C}_0$ and in an element of $\mathcal{C}_1$ (i.e., $\forall C_0 \in \mathcal{C}_0$ $\forall C_1 \in \mathcal{C}_1$ $C_0 \cap C_1 = \{\}$). Let $\mathcal{C} = \mathcal{C}_0 \cup \mathcal{C}_1$.

$\mathcal{O}$ the **observables** is a set of sets of ground atomic formulae. Elements of $\mathcal{O}$ are called **observation alternatives**; elements of observation alternatives are called **atomic observations**. N.B. elements of observation alternatives can unify with the head of rules or can be elements of choice alternatives.

$\pi$ the **observable function**, is a function $\pi : \mathcal{C}_1 \to 2^{\mathcal{O}}$. The idea is that when the agent decides which element of $A \in \mathcal{C}_1$ to choose, it will have 'observed' one atomic observation from each observation alternative in $\pi(A)$. Elements of $\pi(A)$ are the information available to the agent when it has to choose an element of $A$. We assume a 'no forgetting' constraint[2] which means that the element of $\mathcal{C}_1$ are totally ordered and if $C_1 < C_2$ then $C_1 \in \pi(C_2)$ and $\pi(C_1) \subset \pi(C_2)$.

$P_0$ is a function $\cup \mathcal{C}_0 \to [0, 1]$ such that $\forall C \in \mathcal{C}_0$, $\sum_{c \in C} P_0(c) = 1$. I.e., $P_0$ is a probability measure over the alternatives controlled by nature.

$\mathcal{F}$ called the **facts**, is an acyclic logic program such that no atomic choice (in an element of $\mathcal{C}$) unifies with the head of any rule, and such that there is an acyclic ordering [Apt and Bezem, 1991] where every element of every element of $\pi(A)$ is before every element of $A$.

The independent choice logic specifies a particular semantic construction. The semantics is defined in terms of possible worlds. There is a possible world for each selection of one element from each alternative. What follows from these atoms together with $\mathcal{F}$ are true in this possible world.

**Definition 2.3** If $\mathcal{S}$ is a set of sets, a **selector function** on $\mathcal{S}$ is a mapping $\tau : \mathcal{S} \to \cup \mathcal{S}$ such that $\tau(S) \in S$ for all $S \in \mathcal{S}$. The **range** of selector function $\tau$, written $\mathcal{R}(\tau)$ is the set $\{\tau(S) : S \in \mathcal{S}\}$.

**Definition 2.4** Given ICL$_{DT}$ theory $\langle \mathcal{C}_0, \mathcal{C}_1, \mathcal{O}, \pi, P_0, \mathcal{F} \rangle$, let $\mathcal{C} = \mathcal{C}_0 \cup \mathcal{C}_1$. For each selector function $\tau$ on $\mathcal{C}$ there is a **possible world** $w_\tau$. If $f$ is a formula in language $\mathcal{L}_Q$, and $w_\tau$ is a possible world, we write $w_\tau \models f$ (read $f$ is **true in possible world** $w_\tau$) if $\mathcal{F} \cup \mathcal{R}(\tau) \hspace{1pt}\vert\!\sim f$.

The existence and uniqueness of the model follows from the acyclicity of the logic program [Apt and Bezem, 1991].

**Definition 2.5** If $\mathcal{S}$ is a set of sets, the **expansion** of $\mathcal{S}$, written $expansion(\mathcal{S})$ is the set $\{\mathcal{R}(\tau) | \tau$ is a selector function on $\mathcal{S}\}$.

---

[1] Alternatives correspond to 'variables' in decision theory. This terminology is not used here in order to not confuse logical variables (that are allowed as part of the logic program), and random variables. An atomic choice corresponds to an assignment of a value to a variable; the above definition just treats a variable having a particular value as a proposition (not imposing any particular syntax); the syntactic restrictions and the semantic construction ensure that the values of a variable are mutually exclusive and covering, as well as that the variables are unconditionally independent (see [Poole, 1993b])

[2] This constraint can be weakened slightly when the utility can be decomposed into sums [Zhang et al., 1994]

The expansion of $S$ corresponds to the cross product of the elements of $S$ and, when $S$ consists of non-intersecting sets, to the set of minimal hitting sets [Reiter, 1987] of $S$. The set of possible worlds corresponds to the expansion of $C$.

**Definition 2.6** An ICL$_{DT}$ theory is **utility complete** if for each possible world $w_\tau$ there is a unique number $u$ such that $w_\tau \models utility(u)$. The logic program will have rules for $utility(u)$.

**Definition 2.7** An ICL$_{DT}$ theory is **observation inconsistent** if there exists possible world $w_\tau$, and there exists $O \in \mathcal{O}$ with $o_1 \in O, o_2 \in O, o_1 \neq o_2$ such that $w_\tau \models o_1 \wedge o_2$. Otherwise the ICL$_{DT}$ theory is **observation consistent**. An ICL$_{DT}$ theory is **observation complete** if for all possible worlds $w_\tau$, and for all $O \in \mathcal{O}$, there exists $o \in O$ such that $w_\tau \models o$.

The above definitions are to make sure that we can treat the elements of $\mathcal{O}$ as random variables. Unlike elements of $\mathcal{C}$, they are not exclusive and covering 'by definition'. We will always require a theory to be observation consistent, but, when we have negation as failure in the logic [Poole, 1995a] we will not require the theory to be observation complete (there may be an extra, unnamed element of each element of $\mathcal{O}$). Note that observation consistency is not a severe restriction — we can always make $\mathcal{O}$ a set of singleton sets, but then we can't exploit the exclusiveness of observations.

In this paper we assume all our theories are observation consistent and complete.

If an ICL$_{DT}$ theory is observation consistent and observation complete, then for each world $w_\tau$ there is a unique element of $expansion(\mathcal{O})$ that is true in $w_\tau$. Also, for each $C \in \mathcal{C}_1$ there is a unique element of $expansion(\pi(C))$ that is true in $w_\tau$, this element is written here as $\pi_\tau(C)$.

**Definition 2.8** If $\langle \mathcal{C}_0, \mathcal{C}_1, \mathcal{O}, \pi, P_0, \mathcal{F} \rangle$ is an ICL$_{DT}$ theory, then a (pure) **strategy** is a function $\sigma$ on $\mathcal{C}_1$ such that if $C \in \mathcal{C}_1$, $\sigma(C)$ is a function $expansion(\pi(C)) \to C$.

The elements of $expansion(\pi(C))$ are the information available when the decision $C$ is made. In other words a strategy specifies which element of $C$ (for each $C \in \mathcal{C}_1$) to choose ('do') for each of the possible observations.

**Definition 2.9** If ICL$_{DT}$ theory $\langle \mathcal{C}_0, \mathcal{C}_1, \mathcal{O}, \pi, P_0, \mathcal{F} \rangle$ is utility complete, and $\sigma$ is a strategy, then the **expected utility** of strategy $\sigma$ is

$$\varepsilon(\sigma) = \sum_\tau p(\sigma, \tau) \times u(\tau)$$

(summing over all selector functions $\tau$ on $\mathcal{C} = \mathcal{C}_0 \cup \mathcal{C}_1$) where

$$u(\tau) = u \text{ if } w_\tau \models utility(u)$$

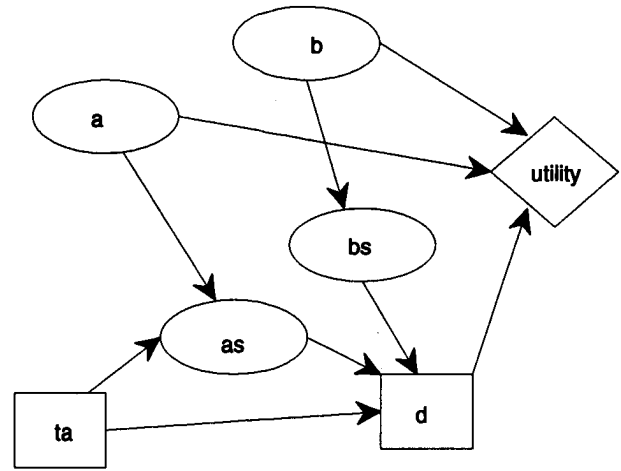

Figure 1: Partially observable, influence diagram

(this is well defined as the theory is utility complete), and

$$p(\sigma, \tau) = \begin{cases} \prod_{C_0 \in \mathcal{C}_0} P_0(\tau(C_0)) & \text{if } \tau(C_1) = \sigma(\pi_\tau(C_1)) \\ & \text{for each } C_1 \in \mathcal{C}_1 \\ 0 & \text{otherwise} \end{cases}$$

$u(\tau)$ is the utility of world $w_\tau$. $p(\sigma, \tau)$ is the probability of world $\tau$ under strategy $\sigma$. For the worlds that could be the result of what the agent chooses (i.e., when the selection function $\tau$ selects the same element of $A$ as does the strategy $\sigma$), the probability is the product of the independent choices of nature. It is easy to show that this induces a probability measure (i.e., for each strategy, the sum of the probabilities of the worlds is 1).

### 2.1 Representing an influence diagram

In this section we axiomatise the influence diagram of Figure 1 in order to demonstrate how the above semantic framework can represent decision problems. This example will also be used to demonstrate our algorithm. In this diagram there is a noisy sensor for $b$, namely $bs$, and a controllable sensor for $a$, namely $as$ (the agent can control which aspect of $a$ it senses).

The problem be represented in our framework as:

$\mathcal{C}_1 = \{\{ta(hi), ta(low)\}, \{d(0), d(1), d(2)\}\}$.
There are two decisions to be made: the agent must choose one of two possible values for $ta$ and one of three possible values for $d$.

$\mathcal{O} = \{\{ta(hi), ta(low)\}, \{as(pos), as(neg)\},$
$\{bs(pos), bs(neg)\}\}$.

$\pi(\{ta(hi), ta(low)\}) = \{\}$.
The agent has no information available when choosing a value for $ta$.

$\pi(\{d(0), d(1), d(2)\}) = \mathcal{O}$.

When choosing a value for $d$ the agent will know values for $ta$, $as$ and $bs$.

$C_0 = \{\{a(low), a(med), a(hi)\}, \{b(pos), b(neg)\},$
$\{false\_pos, true\_neg\}, \{false\_neg, true\_pos\}\}.$
$a$ can have one of three values, $b$ one of two values, the other two alternatives specify the noise of the $bs$ sensor.

Within the facts, we axiomatise how the 'sensors' work.

$bs$ is a noisy sensor of $b$:

$$bs(pos) \leftarrow b(pos) \land true\_pos$$
$$bs(pos) \leftarrow b(neg) \land false\_pos$$
$$bs(neg) \leftarrow b(neg) \land true\_neg$$
$$bs(neg) \leftarrow b(pos) \land false\_neg$$

We can specify the reliability of the sensor as:

$$P_0(false\_pos) = 0.1, P_0(true\_neg) = 0.9$$
$$P_0(false\_neg) = 0.2, P_0(true\_pos) = 0.8$$

$as$ is a sensor which we can control as to whether it detects the high values for $a$ or the low values for $a$ (depending on the value of $ta$):

$$as(pos) \leftarrow ta(hi) \land a(hi)$$
$$as(pos) \leftarrow ta(lo) \land a(lo)$$
$$as(neg) \leftarrow ta(hi) \land a(lo)$$
$$as(neg) \leftarrow a(med)$$
$$as(neg) \leftarrow ta(lo) \land a(hi)$$

We specify the priors on $a$ and $b$ as:

$$P_0(a(low)) = 0.2, P_0(a(med)) = 0.3,$$
$$P_0(a(hi)) = 0.5,$$
$$P_0(b(pos)) = 0.7, P_0(b(neg)) = 0.3$$

Finally the rules specify the utility function.

$$utility(4) \leftarrow d(0)$$
$$utility(10) \leftarrow a(hi) \land d(1)$$
$$utility(3) \leftarrow a(med) \land d(1)$$
$$utility(0) \leftarrow a(lo) \land d(1)$$
$$utility(2) \leftarrow a(hi) \land b(pos) \land d(2)$$
$$utility(5) \leftarrow a(med) \land b(pos) \land d(2)$$
$$utility(9) \leftarrow a(lo) \land b(pos) \land d(2)$$
$$utility(8) \leftarrow b(neg) \land d(2)$$

The above represents the whole decision problem.

Note that the rules for utility and for the probability of $as$ incorporate much more structure than is reflected in the influence diagram.

## 3 So what?

We have presented what seems like quite a complicated semantic construction. The main question arises is "So what; why should anyone be interested?"

First of all this is a language that forms a bridge between the purely logical languages, and the probabilistic and decision theory representations.

If $C_0$ is empty, this is a representation for classical planning that allows for concurrent actions, and uses action completion (in a similar way to how [Reiter, 1991] solves the frame problem). We can axiomatise the world using logic programs. We have, however, a more sophisticated way of handling uncertainty than just disjunction.

Bayesian networks [Pearl, 1988] can be modelled by $C_0$ and $\mathcal{F}$, in the same way that probabilistic Horn abduction [Poole, 1993b] models Bayesian networks. What is added is a richer language for $\mathcal{F}$, with negation as failure and fewer restrictions on the form of the rules [Poole, 1995a], as well as agents with goals [Poole, 1995b].

The language is closely related to influence diagrams [Howard and Matheson, 1981]. Elements of $C_1$ correspond to decision nodes in influence diagrams, with $\pi(A)$ corresponding to the 'parents' of the decision node (these represent information availability when making the decision). The value node is represented as the rules (in $\mathcal{F}$) that imply $utility(u)$ for some $u$.

The main problem considered in this paper is how the representation can be exploited for computational gain.

Influence diagram evaluation procedures can be divided into two classes:

1. Those that do dynamic programming, optimizing the last action first [Shachter, 1986; Cooper, 1988; Shachter and Peot, 1992; Zhang et al., 1994].

2. Those that convert the influence diagram into a decision tree (e.g., [Howard and Matheson, 1981; Qi and Poole, 1995]), and search it using a search method such as *-minimax [Ballard, 1983].

Once it has been realised that efficient Bayesian network algorithms can be used for the probabilistic part of the reasoning [Cooper, 1988; Qi and Poole, 1995], the main cost is in the number of optimizations that needs to be done. For each of the values of the parents of a decision node, one optimization is done. This can be improved in the decision tree methods by not considering those assignments to parents that will have zero probability [Qi and Poole, 1995], but there is still much more that can be done.

The main claim of this paper is that we can exploit the rule-based structure to gain efficiency. We show how the rule structure can be exploited to reduce the number of optimizations (the technique reported here is orthogonal to the idea of removing of impossible conditioning scenarios, so in principle both could be used).

## 4 Abductive characterisation

The structure is exploited by the use of 'explanations'. Instead of reasoning in the space of possible worlds (or in the space of $expansion(\pi(C))$), we reason in the space of explanations. These explanations only make the distinctions needed.

In this section we present the 'abductive' view for the case without negation as failure in the language. There are some interesting issues [Poole, 1995a] that arise in combining this with negation as failure, but these will only complicate this paper.

**Definition 4.1** A set $\kappa$ of atomic choices is **consistent** if there is no alternative $A \in \mathcal{C}$ such that $|A \cap \kappa| > 1$.

**Definition 4.2** A **composite choice** on $\mathcal{K} \subseteq \mathcal{C}$ is a set consisting of exactly one element (atomic choice) from each $C \in \mathcal{K}$.

**Definition 4.3** An **explanation** of ground formula $g$ is a composite choice $\kappa$ on a subset of $\mathcal{C}$ such that $\mathcal{F} \cup \kappa \models g$. A **covering** set of explanations of $g$ is a set of explanations of $g$ such that any explanation of $g$ is a superset of an element of the covering set.

**Definition 4.4** If $G$ is a ground propositional formula, $expl(G)$ is a set of composite choices defined recursively as follows:

$$expl(G) = \begin{cases} \{\} & \text{if } G = true \\ expl(A) \otimes expl(B) & \text{if } G = A \wedge B \\ expl(A) \cup expl(B) & \text{if } G = A \vee B \\ \{\{G\}\} & \text{if } G \in \cup \mathcal{C} \\ \bigcup_i expl(B_i) & \text{if } G \notin \cup \mathcal{C}, \\ & \quad G \leftarrow B_i \in \mathcal{F}' \end{cases}$$

where $\mathcal{K}_1 \otimes \mathcal{K}_2 = \{\kappa_1 \cup \kappa_2 : \kappa_1 \in \mathcal{K}_1, \kappa_2 \in \mathcal{K}_2, consistent(\kappa_1 \cup \kappa_2)\}$. $\mathcal{F}'$ is the set of ground instances of elements of $\mathcal{F}$. $expl$ is well defined as the theory is acyclic.

It can be shown that $expl(g)$ is a covering set of explanations of $g$ (this was essentially proved as the appendix of [Poole, 1993b] and with negation as failure in [Poole, 1995a]) which forms a specification (as a DNF formula of atomic choices) of all of the worlds in which $g$ is true.

Explanations form a concise description of cases (only making distinctions necessary). Sometimes we need to make finer distinctions, for this we need to be able to 'split' composite choices:

**Definition 4.5** If $C = \{c_1, \ldots, c_k\}$ is an alternative and $\kappa$ is a composite choice such that $\kappa \cap C = \{\}$ then the **split** of $\kappa$ on $C$ is the set of composite choices

$$\{\kappa \cup \{c_1\}, \ldots, \kappa \cup \{c_k\}\}$$

It is easy to see that $\kappa$ and a split of $\kappa$ describe the same set of possible worlds. The main use for splitting as described in [Poole, 1995a] is, given a set of composite choices construct a set of mutually incompatible composite choices that describes the same set of possible worlds as the original set. In this paper we show how splitting can be used to construct optimal policies *without* enumerating all information states of a decision.

When we refer to 'the explanations of $g$' are we mean a mutually exclusive (no two explanations are true in any world) and covering set of explanations of $g$, as found for example by $expl$ and either a structure on the rule base to ensure mutual exclusivity (this is the structure achieved by translating a Bayes net into the rules [Poole, 1993b]) or by converting a non-exclusive set of composite choices into an equivalent exclusive set by splitting and subsumption [Poole, 1995a].

## 5 Exploiting the rule structure

In this section we show how to exploit the rule structure. We first give the fully observable case, and show how the rule structure can be used to cut down the case analysis (in a similar way to [Boutilier *et al.*, 1995]). We then show discuss the partially observable case where the observations only give partial information about the state of the world; we then must 'mesh' the cases that make a difference to utility and the cases that can be distinguished by observations.

### 5.1 Fully-observable case

#### 5.1.1 Motivating example

In this section we present an example that not intended to be realistic or meaningful, but demonstrated the algorithm and the some of the savings.

Consider the fully observable influence diagram of Figure 2. Suppose each of the random and decision nodes represent a binary variable. In this influence diagram, if we were to naively apply a dynamic programming procedure, we would optimize the decision $d$ for each of the $2^4 = 16$ values of the parents. Just by looking at this diagram we can see that we do not need to consider the values of $b$ (this is known as 'barren node removal' [Shachter, 1986]). Thus we really only need to consider the $2^3 = 8$ values of the parents of $b$. What is presented in this paper is like the barren node removal, but for specific instances of the parents (e.g., the value of $e$ may be irrelevant for a particular value of $u$).

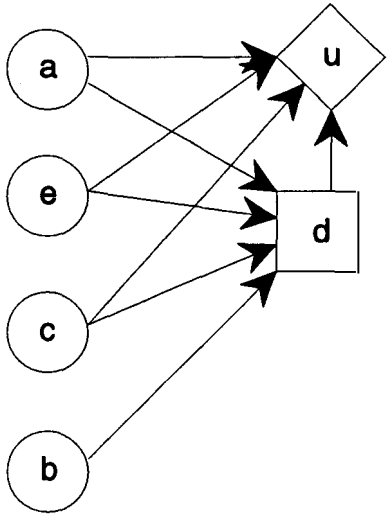

Figure 2: Fully observable, influence diagram

Consider the corresponding ICL$_{DT}$ theory. Here we consider the two values of $a$ to be represented as $a_1$ and $a_2$. $a_1$ is thus the proposition that says that $a$ has one value, and $a_2$ is the proposition that says that $a$ has the other value. The other variables are treated analogously.

$$C_0 = \{\{a_1,a_2\},\{e_1,e_2\},\{c_1,c_2\},\{b_1,b_2\}\}$$
$$C_1 = \{\{d_1,d_2\}\}$$
$$\mathcal{O} = \{\{a_1,a_2\},\{e_1,e_2\},\{c_1,c_2\},\{b_1,b_2\}\}$$
$$\pi(\{d_1,d_2\}) = \{\{a_1,a_2\},\{e_1,e_2\},\{c_1,c_2\},\{b_1,b_2\}\}$$

The value of $P_0$ is irrelevant for the example. Suppose, that the rule-base representation is of the form:

$utility(7) \leftarrow a_1 \wedge d_1$.
$utility(3) \leftarrow a_1 \wedge e_1 \wedge d_2$.
$utility(5) \leftarrow a_1 \wedge e_2 \wedge d_2$.
$utility(4) \leftarrow a_2 \wedge e_1 \wedge c_1 \wedge d_2$.
$utility(5) \leftarrow a_2 \wedge e_1 \wedge c_2 \wedge d_1$.
$utility(6) \leftarrow a_2 \wedge e_1 \wedge c_2 \wedge d_2$.
$utility(7) \leftarrow a_2 \wedge c_1 \wedge d_1$.
$utility(9) \leftarrow a_2 \wedge e_2 \wedge c_1 \wedge d_2$.
$utility(4) \leftarrow a_2 \wedge e_2 \wedge c_2$.

When $a_1$ is true, $c$ and $b$ are irrelevant to the utility. When $a_2 \wedge e_2 \wedge c_2$ is true then the decision is irrelevant. There is even more pruning that can be carried out when we take dominated strategies into account.

### 5.1.2 Finding optimal policies

The fully observable case is where either $C_1$ is empty or there is one decision $d \in C_1$ (the 'last' decision) where $\pi(d) = C - \{d\}$, and when this decision is removed, the remaining theory is fully observable. This case is considered first; the general, partially observable, case will be a modification of this case.

Suppose the last decision is $\{d_1, \cdots, d_k\} \in C_1$.

Consider for each $u$ for which there are rules for $utility(u)$, a covering and exclusive set of explanations of $utility(u)$. The explanations form a partition on the set of possible worlds (each possible world will have exactly one explanation true). For each explanation there are two cases:

1. No $d_i$ is in the explanation. In this case, when this explanation is true, it doesn't matter which decision is made. For example, in the example above $\{a_2, e_2, c_2\}$ is an explanation of a utility that does not involve either $d_1$ or $d_2$. When $a_2 \wedge e_2 \wedge c_2$ is true it doesn't matter which decision is made.

2. For all of the other cases, we treat an explanation as a triple $\langle d_i, \theta, u \rangle$ if $\theta \cup \{d_i\}$ is an explanation for $utility(u)$. If this is the case then the algorithm of Figure 3 will compute the optimal policy. The algorithm repeatedly removes dominated explanations and splits explanations where finer distinctions are needed.

   At the end of the algorithm, the resulting explanations corresponds to an optimal policy with $\langle d_i, \theta, u \rangle$ meaning "do $d_i$ if $\theta$ is true, and $u$ will be the utility". If more than one of the $\theta_i$ is true, it doesn't matter which action is chosen (either the actions will be the same or the utilities will be the same). The $\theta_i$ will cover all of the cases not covered in case 1 above.

The worst case of this algorithm occurs when we have to split on all alternatives — this is the same as enumerating all states of the observables.

In our example above, for the case where the decision matters (i.e., for all cases except when $a_2 \wedge e_2 \wedge c_2$) we have the following explanations:

$\langle d_1, \{a_1\}, 7\rangle$ (1)
$\langle d_2, \{a_1, e_1\}, 3\rangle$ (2)
$\langle d_2, \{a_1, e_2\}, 5\rangle$ (3)
$\langle d_2, \{a_2, e_1, c_1\}, 4\rangle$ (4)
$\langle d_1, \{a_2, e_1, c_2\}, 5\rangle$ (5)
$\langle d_2, \{a_2, e_1, c_2\}, 6\rangle$ (6)
$\langle d_1, \{a_2, c_1\}, 7\rangle$ (7)
$\langle d_2, \{a_2, e_2, c_1\}, 9\rangle$ (8)

Explanations (2) and (3) are dominated by (1) and can be removed. (5) is dominated by (6) and can be removed. (4) is dominated by (7) and can be removed. We split (7) on $\{e_1, e_2\}$, resulting in explanations $\langle d_1, \{a_2, e_1, c_1\}, 7\rangle$ and $\langle d_1, \{a_2, e_2, c_1\}, 7\rangle$, the latter of which can be pruned as it is dominated by (8).

---

**procedure** $optimize(S)$:

input: set $S$ of tuples of the form $\langle d_i, \theta, u \rangle$, such that whenever composite choice $\theta$ is true, decision $d_i$ has utility $u$.

output: set $S$ of tuples of the form $\langle d_i, \theta, u \rangle$, such that whenever $\theta$ is true, decision $d_i$ has utility $u$, and $d_i$ is an optimal decision when $\theta$ is true.

1. Remove all dominated explanations from $S$. If we have two elements $\langle d_i, \theta_i, u_i \rangle$ and $\langle d_j, \theta_j, u_j \rangle$ in $S$ where $u_i \leq u_j$ and $\theta_j \subseteq \theta_i$, then remove $\langle d_i, \theta_i, u_i \rangle$. If $u_i = u_j$ and $\theta_j = \theta_i$ then remove either one. [Whenever $\theta_i$ is true, we know that $d_j$ is better than $d_i$, and so we don't need to consider $d_i$.]

2. If there are no dominated elements, and if there are two elements $\langle d_i, \theta_i, v_i \rangle \in S$ and $\langle d_j, \theta_j, v_j \rangle \in S$ such that $\theta_i \cup \theta_j$ is consistent, $d_i \neq d_j$ and $v_i < v_j$ do the following: Select $o \in \theta_j - \theta_i$. Suppose $o$ is in observation alternative $O$. Replace $\langle d_i, \theta_i, v_i \rangle$ in $S$ by the split of $\langle d_i, \theta_i, v_i \rangle$ on $O$. Go to step 1.

3. If neither case 1 nor case 2 is applicable, return $S$.

Figure 3: Finding optimal policies from observations

---

The resulting explanations are:

$\langle d_1, \{a_1\}, 7 \rangle$
$\langle d_2, \{a_2, e_1, c_2\}, 6 \rangle$
$\langle d_1, \{a_2, e_1, c_1\}, 7 \rangle$
$\langle d_2, \{a_2, e_2, c_1\}, 9 \rangle$

This can be interpreted as an optimal policy, which says that when $a_1$ is true do $d_1$, when $a_2 \wedge e_1 \wedge c_2$ is true do $d_2$, etc. When none of the cases occur (i.e., when $a_2 \wedge e_2 \wedge c_2$ is observed) it doesn't matter which action is taken.

### 5.2 Partially Observable Single Decision

The partially observable single decision case consists of 4 steps:

1. finding explanations of $utility(U)$ for each value $U$ and explanations for the observables, using $expl$;

2. repeated removing of dominated explanations and case analysis on relevant observations;

3. computed expected utilities for the relevant cases of observations and

4. using the *optimize* algorithm of Figure 3 to generate an optimal strategy.

Suppose we have the last decision $d = \{d_1, \ldots, d_k\} \in \mathcal{C}_1$. Dynamic programming would tell us that we have to consider each case of $expansion(\pi(d))$ (the set of all value assignments to the 'parents' of $d$) — there are exponentially many of these (in the size of $\pi(d)$). We would like to consider each observable in $\pi(d)$ separately, to see when it is relevant to the decision being made. This is done by the use of œ:

**Definition 5.1** If $O \in \pi(d)$ let $œ(O) = \{\langle o, expl(o) \rangle : o \in O\}$. $œ(O)$ is a representation for the explanations of the possible observations in $O$. We assume that for each $O \in \pi(d)$, that the function $œ(O)$ is computed once and stored.

**Example 5.2** Consider the example of Section 2.1.

For the last decision, namely $\{d(0), d(1), d(2)\}$, and for each 'observable', we determine the œ function which tells us what it is that the sensors detect:

$œ(\{ta(hi), ta(low)\})$
$= \{\langle ta(hi), \{\{ta(hi)\}\} \rangle, \langle ta(lo), \{\{ta(lo)\}\} \rangle\}$

$œ(\{as(pos), as(neg)\})$
$= \{\langle as(pos), \{\{ta(hi), a(hi)\}, \{ta(lo), a(lo)\}\} \rangle,$
$\quad \langle as(neg), \{\{ta(hi), a(lo)\}, \{a(med)\},$
$\quad\quad \{ta(lo), a(hi)\}\} \rangle\}$

$œ(\{bs(pos), bs(neg)\})$
$= \{\langle bs(pos), \{\{b(pos), true\_pos\},$
$\quad\quad \{b(neg), false\_pos\}\} \rangle,$
$\quad \langle bs(neg), \{\{b(neg), true\_neg\},$
$\quad\quad \{b(pos), false\_neg\}\} \rangle\}$

These are stored, and are combined with the explanations of different utilities in order to determine the relevant cases.

In general we have to check correlations between the observations and the cases where one decision is better than the other. The fully observable case showed the idea of to how to isolate the cases where the one decision is better than another.

**Definition 5.3** Tuple $\langle d_i, \theta, \mathcal{K}, u \rangle$, where $\theta$ is a composite observation and $u$ is a number is a **pre-policy** if $\mathcal{K}$ is a set of covering explanations of $\theta \wedge utility(u)$ which contains $d_i$. Set $E_0$ of pre-policies is **covering** if for every world $w_\tau$ there is a unique $\langle d_i, \theta, \mathcal{K}, u \rangle \in E_0$ such that $d$ is true in $w_\tau$ and one explanation in $\mathcal{K}$ is true in $w_\tau$.

The algorithm works by maintaining a set of covering pre-policies. These can be set up using:

**Lemma 5.4** The set $\{\langle d_i, \{\}, \{\mathcal{K} \in expl(utility(u)) : d_i \in \mathcal{K}\}, u \rangle : d_i \in d\}$ is a covering set of pre-policies. This can be easily computed by generating $expl(utility(u))$ for each $u$ for which there are rules.

**Example 5.5** Continuing our example, we create the pre-policies for different utilities, some of which are:

$$\langle d(0), \{\}, \{\{d(0)\}\}, 4\rangle \quad (9)$$
$$\langle d(1), \{\}, \{\{d(1), a(hi)\}\}, 10\rangle \quad (10)$$
$$\langle d(1), \{\}, \{\{d(1), a(med)\}\}, 3\rangle$$

These specify the distinctions that are important to determine utility.

A basic step is to split pre-policies based on the different possible values of an observable (i.e., we consider each of the cases for the values of the observable):

**Lemma 5.6** If $\langle d, \theta, \mathcal{K}, u\rangle$ is a pre-policy, and $o \in \pi(d)$ then for all $\langle c, \mathcal{K}'\rangle \in \text{œ}(o)$, $\langle d, \theta \cup \{c\}, \mathcal{K} \otimes \mathcal{K}', u\rangle$ is a pre-policy.

We only want to do case analysis with respect to an observation if it is relevant. The notion of 'autonomous' gives a syntactic criteria for determining if an observation is relevant:

**Definition 5.7** If $\mathcal{K}_1$ and $\mathcal{K}_2$ are sets of composite choices then $\mathcal{K}_1$ and $\mathcal{K}_2$ are **autonomous** if $\forall \kappa_1 \in \mathcal{K}_1 \forall \kappa_2 \in \mathcal{K}_2 \forall c_1 \in \kappa_1 \forall c_2 \in \kappa_2 \nexists A \in \mathcal{C} \{c_1, c_2\} \subseteq A$. Thus they are autonomous if they involve different alternatives.

The following lemma can be easily proved:

**Lemma 5.8** If the set of explanations of $g_1$ and the set of explanations of $g_2$ are autonomous then $g_1$ and $g_2$ are independent.

We can stop expanding on observations when all other observations are irrelevant.

**Definition 5.9** Pre-policy $\langle d_i, \theta_1, \kappa_1, u\rangle$ is **observation full** if for every $\theta_2 \in \pi(d)$ either $\theta_2 \cap \theta_1 \neq \{\}$ or for all $\langle c, \kappa_2\rangle \in \text{œ}(\theta_2)$, $\kappa_2$ and $\kappa_1$ are autonomous.

If pre-policy $\langle d, \theta, \kappa_1, u\rangle$ is observation full, then the other observations are irrelevant to decision $d$ in the context of observation $\theta$.

**Example 5.10** Continuing example 5.5, partial explanation (9) is observation full: for action $d(0)$ all observations are irrelevant as far as the utility is concerned.

Partial explanation (10) is autonomous of $\text{œ}(\{ta(hi), ta(low)\})$ and $\text{œ}(\{bs(pos), bs(neg)\})$, but is not autonomous of $\text{œ}(\{as(pos), as(neg)\})$.

Figure 4 given a procedure for expanding a set of pre-policies to an equivalent set that is observation-full. In the worst case, the set produced will contain one element for each element of $d$ and each element of $expansion(\pi(d))$. In many cases this will be much smaller. This algorithm

---

**procedure** $expand(S)$:

input: set $S$ of pre-policies.

output: set of observation full pre-policies.

1. Select $\langle d_i, \theta, \mathcal{K}, u\rangle \in S$ and $O \in \pi(d)$ such that $O \cap \theta = \{\}$ and there is one $\langle c, \mathcal{K}'\rangle \in \text{œ}(O)$ where $\mathcal{K}'$ and $\mathcal{K}$ are not autonomous.
   $S := S - \{\langle d_i, \theta, \mathcal{K}, u\rangle\} \cup \{\langle d_i, \theta \cup \{c\}, \mathcal{K} \otimes \mathcal{K}', u\rangle : \langle c, \mathcal{K}'\rangle \in \text{œ}(O)\}$.
   Go to step 1.

2. If there are no choices in case 1, return $S$.

Figure 4: Expanding observations to cover all potentially relevant cases

---

contains a selection — the algorithm will be correct no matter which elements (that satisfy the conditions) are selected. Different selections may change the size of the resulting set (e.g., of one observation gives more information than another, this observation should be selected first). Also note that the case analysis we do for the observations is not symmetric — one observation may only be relevant for particular values of other observations.

**Example 5.11** Partial explanation (10) needs to be combined with $\text{œ}(\{as(pos), as(neg)\})$ resulting in:

$$\langle d(1), \{as(pos)\}, \{\{d(1), a(hi), ta(hi)\}\}, 10\rangle$$
$$\langle d(1), \{as(neg)\}, \{\{d(1), a(hi), ta(lo)\}\}, 10\rangle$$

These can the be combined with $\text{œ}(\{ta(hi), ta(low)\})$ producing:

$$\langle d(1), \{as(pos), ta(hi)\}, \{\{d(1), a(hi), ta(hi)\}\}, 10\rangle$$
$$\langle d(1), \{as(neg), ta(lo)\}, \{\{d(1), a(hi), ta(lo)\}\}, 10\rangle$$

Which are then observation full. For decision $d(1)$ value of $bs$ is irrelevant.

Once we have an observation full set of pre-policies, we can then compute expected values (of the utility given decisions and observations), using the algorithm of Figure 5. Computing expected values is complicated by the fact that for a $d_i$ the pre-policies involving different utility values may involve a different split on the observations. This algorithm computes expected utilities for combinations of values of observables, splitting cases when necessary. This procedure treats the expectation calculation as one big sum; in any real implementation we would use some of the more efficient Bayesian network algorithms (such as clique-tree propagation).

Given the expected utilities of the observables, we essentially have the fully observable case from which we can then use *optimize* of Figure 3 to derive the optimal policy.

**procedure** $expectation(S)$:
input: set of observation full pre-policies.
output: set of tuples of the form $\langle d_i, \mathcal{K}, v \rangle$, such that whenever $\mathcal{K}$ is observed, decision $d_i$ has (expected) utility $v$.

while $\exists \langle d_i, \theta, \mathcal{K}, u \rangle \in S, \langle d_i, \theta', \mathcal{K}', u' \rangle \in S$
such that $consistent(\theta \cup \theta')$ and $\theta \not\subseteq \theta'$
select $\omega \in \theta - \theta'$
let $\Omega$ be the element of $\mathcal{O}$ such that $\omega \in \Omega$
replace $\langle d_i, \theta', \mathcal{K}', u' \rangle$ in $S$ by
the split of $\langle d_i, \theta', \mathcal{K}', u' \rangle$ on $\Omega$
end while;
Let

$$\varepsilon(d_i, \theta, S) = \frac{\sum_{\langle \mathcal{K}, u \rangle : \langle d_i, \theta, \mathcal{K}, u \rangle \in S} P(\mathcal{K}) \times u}{\sum_{\langle \mathcal{K}, u \rangle : \langle d_i, \theta, \mathcal{K}, u \rangle \in S} P(\mathcal{K})}$$

$expectation(S)$
$= \{\langle d_i, \theta, \varepsilon(d_i, \theta, S) \rangle : \exists \mathcal{K} \exists u \, \langle d_i, \theta, \mathcal{K}, u \rangle \in S\}$

Figure 5: Computing Expectations

The general algorithm is now to compute the policy via:

$S_0 := \{\langle d_i, \{\}, \{\mathcal{K} \in expl(utility(u)) : d_i \in \mathcal{K}\}, u \rangle$
$\qquad : d_i \in d\};$
$S_1 := optimize(expectation(expand(S_0)))$

Then $S_1$ corresponds to the optimal policy.

### 5.3 Multiple Decisions

We assume that there is a 'last' decision $d = \{d_1, \ldots, d_k\} \in \mathcal{C}_1$, such that all other decisions that are part of an explanation of $utility$ which contain $d$ are in $\pi(d)$ (i.e., are 'observable'). If there is no such decision, then we cannot optimize the decisions one at a time [Zhang et al., 1994].

The idea of the algorithm for multiple decisions is the standard one: we solve the last decision and either replace it with a deterministic function corresponding to the policy (by adding the corresponding rules to $\mathcal{F}$), or by replacing the rules for utility by new rules that give the expected utility for the optimal policy [Zhang et al., 1994].

### 5.4 Refinements

There are many refinements that can be given to the above procedure. A few are noteworthy:

1. We want to do subsumption as early as possible. Subsumption can be made as early as the expand procedure. Note that, for those cases where all alternatives have been subsumed, neither the $expectation$ procedure nor the $optimize$ procedure need to do any splitting.

2. We really want to compute the explanations and the other procedures in a lazy fashion — only expanding enough to see what can be pruned. We want to prune early and prune often!

3. Although we have specified $expl$ here as an abstract procedure, it can be computed top-down (as in [Poole, 1993a]), bottom-up (as in an ATMS), and we are also exploring exploiting structure in a rule-based version of clique-tree propagation.

## 6 Conclusion

This paper presented one step in a combination of logic and probability.

This paper has proposed a mechanism for reducing the case analysis of dynamic programming. We have exploited the rule structure of the ICL$_{\text{DT}}$ in order to determine the cases where some observations are irrelevant.

The use of rule structure (called 'tree-structure' there) for Markov decision processes has been explored by Boutilier and colleagues [Boutilier et al., 1995]. Their algorithm is similar to the 'fully observable' case of section 5.1. This paper expands on this to only consider appropriate groupings of observations.

Smith et. al [Smith et al., 1993] have explored the use of tree-like definitions for the conditional probability tables in influence diagrams. The difference to this work is that we only have rules plus independent choices — the influence diagram is just one of the representations we can represent. The algorithms are also very different, with Smith et. al. using a variant of Shachter's algorithm.

This is a part of a project to create a mix of logic and decision theory, where we can exploit as much of the structure as possible to gain efficiencies. This paper has only scratched the surface of the issues. Currently under development (or under consideration) are rule-based variants (that can exploit the propositional independence inherent in a rule base) of common probabilistic algorithms for MDPs [Boutilier et al., 1995], influence diagrams (this paper), POMDPs and even a rule-based version of clique-tree propagation.

The representation used here is also of interest in its own right in providing logical variables that can be used for dynamic construction of decision networks (as in [Poole, 1993b]), and can be extended into multiple agents [Poole, 1995b].

**Acknowledgements**

This work was supported by Institute for Robotics and Intelligent Systems, Project IC-7 and Natural Sciences and Engineering Research Council of Canada Operating Grant OGP0044121. Thanks to Craig Boutilier for valuable discussions and for comments on earlier versions of this paper.